\documentclass[letterpaper, 10 pt, conference]{ieeeconf}  

\IEEEoverridecommandlockouts                              

\usepackage{multirow}
\usepackage{graphicx}

\title{\LARGE \bf
Distributed Learning for UAV Swarms
}

\author{Chen Hu$^{1}$, Hanchi Ren$^{1}$, Jingjing Deng$^{2}$, Xianghua Xie$^{1}$*
\thanks{$^{1}$Chen Hu, Hanchi Ren, and Xianghua Xie are with the Computer Science Department, Faculty of Science and Engineering, Swansea University, UK}
\thanks{$^{2}$Jingjing Deng is with Department of Computer Science, Durham University, UK}%
\thanks{$^{*}$Corresponding Author: Xianghua Xie ({\tt\small x.xie@swansea.ac.uk})}
}

\begin{document}

\maketitle
\thispagestyle{empty}
\pagestyle{empty}

\begin{abstract}
Unmanned Aerial Vehicle (UAV) swarms are increasingly deployed in dynamic, data-rich environments for applications such as environmental monitoring and surveillance. These scenarios demand efficient data processing while maintaining privacy and security, making Federated Learning (FL) a promising solution. FL allows UAVs to collaboratively train global models without sharing raw data, but challenges arise due to the non-Independent and Identically Distributed (non-IID) nature of the data collected by UAVs. In this study, we show an integration of the state-of-the-art FL methods to UAV Swarm application and invetigate the performance of multiple aggregation methods (namely FedAvg, FedProx, FedOpt, and MOON) with a particular focus on tackling non-IID on a variety of datasets, specifically MNIST for baseline performance, CIFAR10 for natural object classification, EuroSAT for environment monitoring, and CelebA for surveillance. These algorithms were selected to cover improved techniques on both client-side updates and global aggregation. Results show that while all algorithms perform comparably on IID data, their performance deteriorates significantly under non-IID conditions. FedProx demonstrated the most stable overall performance, emphasising the importance of regularising local updates in non-IID environments to mitigate drastic deviations in local models.

\end{abstract}

\section{INTRODUCTION}

Unmanned aerial vehicle (UAV) swarms are being adopted in a wide array of real-world applications due to their ease of deployment and flexible manoeuvrability. These applications range from environmental sensing and surveillance to communication and rescue missions \cite{uav-comm, uav-search, uav-civil}. UAV swarms in the real world often require UAVs to operate in dynamic and distributed environments, where they collect diverse data stored on-board. For instance, in a surveillance scenario, a UAV swarm may capture images at various locations for different targets to assess the situation. Such applications benefit from UAVs working together to analyse their data and generate insights, but the challenge of processing this data while maintaining security and privacy remains a critical concern.

\begin{figure}[h]
\centering
\includegraphics[width=1.0\linewidth]{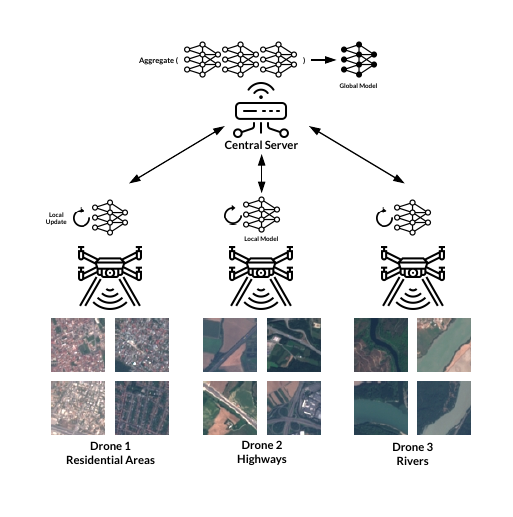}
\caption{Federated Learning Framework for UAV Swarms. Each UAV trains a local model on collected data and sends only the model to a central server. The server aggregates the models and updates a global model, which is then shared with the UAVs for further training. In real-world applications, the data collected by different UAVs is typically heterogeneous.}
\label{fed}
\end{figure}

Federated learning (FL) offers a promising solution for enabling UAV swarms to collaboratively train a global machine learning model without the need to exchange or upload the raw data. Withing the FL framework, each UAV trains a local model on its own data and shares only the model with a central server. This decentralised paradigm allows UAV swarms to analyse the data they collect locally, thereby enhancing privacy and security by keeping sensitive information confined to each UAV. Figure \ref{fed} shows an example framework for deploying FL in a UAV swarm for environmental monitoring.

However, real-world deployments of UAV swarms often result in heterogeneity in the data collected by each UAV, leading to non-identically distributed (non-IID) data. For instance, UAVs might capture vastly different images based on their geographic location, altitude, or specific mission, causing great variation in the data across UAV nodes. This non-IID nature presents a unique challenge for federated learning, as local models trained on non-IID data can generate imbalanced or even conflicting updates. These inconsistent local updates can cause difficulties in global model convergence, as the aggregation of contradictory model updates may slow down or diverge the learning process.

Current research on applying FL in UAV swarms has primarily focused on leveraging FL for tasks such as environmental sensing \cite{uav-air-0, uav-air-1, uav-sense}, or optimising communication and resource allocation strategies to enhance the efficiency of FL in such networks \cite{uav-opt-0, uav-opt-1, uav-opt-2, uav-opt-3}. While these efforts are valuable, there has been limited study on how federated learning algorithms handle the critical issue of data heterogeneity in UAV swarms. Only a few studies \cite{uav-imba} have specifically addressed the challenge of non-IID data, leaving a gap in the literature when it comes to effectively deploying FL in real-world UAV systems that must contend with non-IID data distributions.

In this study, we carry out integrations of ,ultiple federated learning algorithms under both IID and non-IID settings to investigate their robustness to heterogeneous data distributions. We conduct our experiments using four diverse datasets: MNIST \cite{mnist}, CIFAR10 \cite{cifar}, EuroSAT \cite{eurosat}, and CelebA \cite{celeba}. MNIST serves as a baseline to assess performance on a relatively simple classification task. CIFAR10 is used to simulate a more complex object classification task, reflecting the kinds of challenges UAV swarms might face in real-world visual recognition scenarios. EuroSAT is employed to mimic the use of UAV swarms for monitoring land use and land cover, providing an example of UAV applications in environmental and geographic data collection. Finally, CelebA is used to simulate the collaborative learning of a face recognition system by UAVs, which could be relevant for surveillance or search-and-rescue operations. Through this evaluation across different datasets and scenarios, we aim to shed light on how various FL algorithms handle data heterogeneity and assess their potential for deployment in real-world UAV swarm applications.

\section{Methods}

In the section, we introduce the federated learning algorithms and the neural network architectures used in this study. We explore FedAvg \cite{fedavg}, FedProx \cite{fedprox}, FedOpt \cite{fedopt} and MOON \cite{moon}, each of which employs distinct strategies to handle the distributed learning task. FL algorithms generally involve two stages of updates, namely local updates on client nodes and global aggregation on the server side. These two stages are repeated iteratively until the global model is converged. When deploying FL to UAV swarms, we assume that each UAV has finished data collection and we focus on training the models with FL. Among the algorithms we consider, FedAvg serves as a baseline, providing straightforward model aggregation through averaging local updates. FedProx extends FedAvg by adding a proximal term to mitigate the impact of non-IID data. FedOpt enhances the global optimisation process by introducing server-side optimisation techniques. MOON encourages local learned representations to diversify while bounding these representations to the global representations. We summarize algorithms discussed in this section in Table \ref{summary}.

\begin{table}[ht]
\caption{Summary of Evaluated Federated Learning Algorithms}
\centering
\begin{tabular}{lll}
\hline
Method  & Applicable Stage   & Approach                                                                            \\ \hline
FedAvg  & -                  & -                                                                                   \\
FedProx & local update       & adding local proximal term                                                          \\
FedOpt  & server aggregation & \begin{tabular}[c]{@{}l@{}}employing momentum \\ during server updates\end{tabular} \\
MOON    & local update       & \begin{tabular}[c]{@{}l@{}}contrasting local-global \\ representations\end{tabular} \\ \hline
\end{tabular}
\\ FedAvg serves as the baseline for other FL algorithms.
\label{summary}
\end{table}

\noindent \textbf{FedAvg} Federated Averaging (FedAvg) is one of most commonly used methods in federated learning. The central server first initialises the global model parameters $w_g$ at the start of training. This global model is shared with all UAVs in the swarm. Next, each UAV receives the global model $w_g$ and performs local training on its collected data using gradient descent,

\begin{equation}
    w_l^t = w_g^t, w_l^{t+1} = w_l^t - \eta \nabla_l L(w_l^t)
\end{equation}
\noindent where $w_l^t$ is the local model for the $t$-th training round, $\eta_l$ is the local learning rate, and $L$ is the local training loss. Each UAV performs multiple update steps, however, the number of local training steps are typically much fewer than in traditional centralised learning.

After the local training stage, UAVs that participate the current training round send their local models $w_l$ to the central server. The central server aggregates local models by averaging their weights,

\begin{equation}
    w_g^{t+1} = \sum_l \frac{n_l}{N}w_l^t
\end{equation}
where $n_l$ is the number of training samples on the $l$-th UAV node, $N$ is the total number of training samples across UAVs that participate the current training round. In another word, the server averages local models weighted by the amount of data on each UAV, ensuring that UAVs with more data samples have a greater influence on the global model. The process of local training and global aggregation is repeated until the global model converges to a desired performance level.

\noindent \textbf{FedProx} FL deployed in UAV swarms inherently faces challenges posed by the non-IID nature of data across distributed UAV nodes. Algorithms such as FedAvg assume that the data distribution across participating computation nodes is homogeneous and all devices contribute equally during training. However, this assumption is unlikely to hold in real-world applications as UAVs may collect heterogeneous data due to their location and environmental differences. FedProx is an extension of FedAvg, emphasising on addressing the non-IID data challenges in FL. FedProx introduces a proximal term in the local loss function of each client. This term acts as the regularisation and constrains local updates to be closer to the global model. The added proximal term also helps to mitigate the impact of non-IID data by preventing large local model updates that deviate from the global model. The local update for FedProx is formulated as, 

\begin{equation}
    w_l^{t+1} = w_l^t - \eta_l \nabla_l [L(w_l^t) + \frac{\mu}{2} \|w_g^t - w_l^t\|^2 ]
\end{equation}
\noindent where $\mu$ is the weight for the proximal term that controls the degree of regularisation. The global aggregate stage for FedProx is the identical to that of FedAvg.

\noindent \textbf{FedOpt} In contrast to FedProx that regularises client local updates, FedOpt aims to improve the optimisation of the global model by incorporating more sophisticated optimisation on the server side, which includes the use of momentum and adaptive learning rate. Instead of directly sending model weights to the central server, FedOpt requires individual UAVs to upload the model difference between the updated local model and current global model. These model differences are treated as gradients and used to compute the momentum term for server updates. The server side update in FedOpt can be formulated as,

\begin{equation}
    w_g^{t+1} = w_g^t + \eta_g \frac{m_t}{\sqrt{v_t}+\tau}
\end{equation}
where $\eta_g$ is the learning rate on the server side, $m_t$ is the moving average of local model differences, $v_t$ is the momentum term computed from average local differences using Adam \cite{adam} or Adagrad \cite{adagrad}, and $\tau$ is a hyperparameter that controls the adaptivity of server side updates. 

\noindent \textbf{MOON} To address the issue of non-IID data in FL, MOON introduces contrastive learning to local updates. In addition to each client's local learning objective, MOON further encourages the local final representations (input to the classifier head) to be close to the representations extracted from current global model, while pushing the distances between current local representations and representations from the previous local model. That is, MOON prevents local representations to deviate too much from the global representation, and diversify local representations over time. The local update in MOON is formulated as

\begin{equation}
    w_l^{t+1} = w_l^t - \eta_l \nabla_l [L(w_l^t) - \mu \log \frac{\exp(f_l^t f_g^t)}{\exp(f_l^t f_g^t) + \exp (f_l^t f_l^{t - 1})}]
\end{equation}
\noindent the negative log term above is the the NT-Xent loss \cite{ntloss}, $f_l^t$ is the representation from the local model at the $t$-th round, $f_g^t$ is the representation from the global model at the $t$-th round, and $\mu$ is the hyperparameter that controls the weight of the contrastive term. The global model aggregation stage of MOON is also identical to that of FedAvg.

\begin{table*}[ht]
\caption{Summary of Experiment Configurations}
\centering
\begin{tabular}{cccccc}
\hline
Dataset & Domain             & Number of Samples & Number of Classes & Number of Clients & Training Rounds\\ \hline
MNIST   & handwritten digits & 60,000            & 10                & 100 & 200            \\
CIFAR10 & natural objects     & 60,000            & 10                & 100 & 500               \\
EuroSAT & remote sensing     & 27,000            & 10                & 100 & 300               \\
CelebA  & human faces        & 6,038             & 200               & 50 & 500                \\ \hline
\end{tabular}
\label{dataset}
\end{table*}

\subsection{Model Architectures and Loss Functions}
To better accommodate the resource constraints of UAV swarms, we primarily utilise lightweight neural network architectures in this study. Specifically, for experiments on MNIST, we employ a simple feedforward neural network composed of three fully connected layers, with the hidden layer consisting of 200 neurons.

\begin{figure}[ht]
\centering
\includegraphics[width=1\linewidth]{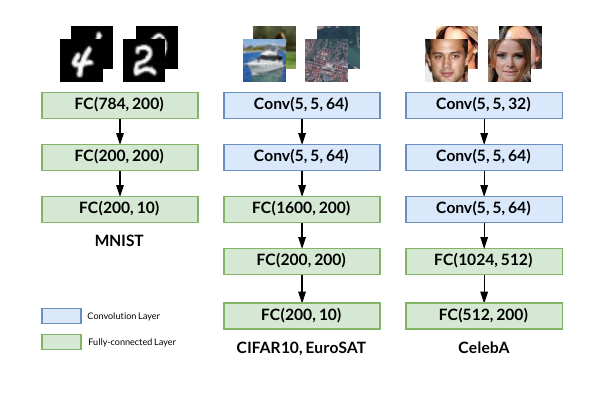}
\caption{Neural Network Architectures Used for Experiments. The activation function and pooling layers are ignored in this figure.}
\label{model}
\end{figure}

For the experiments on CIFAR-10, EuroSAT, and CelebA, we use convolutional neural networks (CNNs) to better capture spatial features in the image data. The CNNs for CIFAR-10 and EuroSAT consist of two convolutional layers with a kernel size of 5$\times$5, followed by two fully connected layers. The CNN architecture for CelebA includes an additional convolutional layer due to the larger input image size. The architectures of these models are shown in Figure \ref{model}.

Regarding the local loss functions, we use the cross-entropy loss across all datasets, as the local learning tasks are all multi-class classification. Additionally, we include regularisation terms as needed, depending on the specific federated learning algorithm introduced in the previous section. In the case of the face recognition task on CelebA, we also experiment with the ArcFace margin loss \cite{arcface}, which is designed to enhance the discriminative capability of the model by increasing the margin between learned face representations. The results of these experiments are discussed in detail in the experimental section.

\begin{table*}[ht]
\centering
\caption{Accuracy Test Performance on Evaluated Datasets}
\begin{tabular}{ccccccc}
\hline
        & \multicolumn{2}{c}{MNIST} & \multicolumn{2}{c}{CIFAR10} & \multicolumn{2}{c}{EuroSAT} \\ \cline{2-7} 
Method & IID & Non-IID & IID & Non-IID & IID & Non-IID \\ \hline
FedAvg & 97.83 & 96.12 & 71.03 & 48.62 & 85.14 & 71.18 \\
FedProx & \textbf{97.99} & \textbf{96.34} & 71.12 & 46.77 & \textbf{85.16} & \textbf{75.67} \\
FedOpt & 97.75 & 96.26 & 68.78 & \textbf{57.59} & 80.56 & 73.53 \\
MOON & 96.12 & 95.93 & \textbf{71.98} & 42.97 & 84.38 & 67.39 \\ \hline
\end{tabular}
\\ \%-accuracy of federated learning algorithms evaluated in settings with identical or non-identical data distributions. 
\\ The higher the accuracy the better the performance. The best results are highlighted \textbf{in bold}.
\label{classify}
\end{table*}

\section{Experiments}
\subsection{Datasets and Evaluation Metrics}
To evaluate the federated learning algorithms discussed in the method section, we employ the MNIST, CIFAR10, EuroSAT, and CelebA datasets. We summary the details of each dataset and the number of simulated UAV clients in Table \ref{dataset}.

The MNIST dataset consists of 60,000 grayscale images, each of size 28$\times$28, representing handwritten digits ranging from 0 to 9. This dataset is a standard benchmark in image classification. CIFAR10 contains 60,000 colour images, each of size 32$\times$32, distributed across 10 distinct classes, including airplanes, automobiles, birds, cats, deer, dogs, frogs, horses, ships, and trucks. EuroSAT is a remote sensing image dataset composed of 27,000 images from 10 different land-use and land-cover classes, such as agricultural areas, residential zones, forests, lakes, and industrial sites. Each image was originally collected at 64$\times$64 resolution, but for this study, we resize the images to 32$\times$32. CelebA is a dataset of human face images that includes over 200,000 samples covering 10,177 subjects. For this study, we focus on a subset of the top 200 subjects with the most samples, yielding a dataset with 6,038 face images. We employ the Multi-task Cascaded Convolutional Networks (MTCNN) \cite{mtcnn} to detect and crop the faces from the raw images, resizing them to 64$\times$64 pixels.

To simulate UAV swarms, we distribute the MNIST, CIFAR10, and EuroSAT datasets across 100 clients, representing individual UAVs in the federated learning framework. The CelebA subset, which has fewer samples, is distributed to 50 clients. For all datasets, we maintain an 80/20 split, with 80\% of the data used for training and 20\% for testing. We train for 200 rounds on the MNIST, 300 rounds on EuroSAT, and 500 rounds on both CIFAR10 and CelebA.

We evaluate federated learning under both identical and non-identical data distribution settings. In the identical data distribution setting, training samples are randomly and evenly distributed among all UAV clients, ensuring that each client has a balanced dataset covering all classes. This setup represents an ideal scenario. To simulate more realistic and heterogeneous data distribution across clients, we adopt the pathological non-IID setting described in the original FedAvg paper. In this setting, each client only has access to a subset of the classes. We limit the number of classes on each client to 4. This non-IID scenario mimics the heterogeneity of data that UAVs may collect in the real world.

To measure the performance of the federated learning algorithms, we primarily use classification accuracy on the test data across all datasets. For MNIST, CIFAR10, and EuroSAT, accuracy reflects the model’s ability to correctly classify samples from the respective datasets.

In addition to standard classification tasks, we conduct a specialized evaluation for face recognition models trained on CelebA. Following common face recognition protocols, we perform the LFW (Labeled Faces in the Wild) test \cite{lfw}, which evaluates the model's ability to distinguish between pairs of face images. Specifically, the model is tasked with comparing image pairs to determine whether they belong to the same individual. This is done by feeding the image pairs into the model and comparing the similarity between the facial representations extracted from each image. The model is expected to generate similar features for images of the same person and distinct representations for different individuals.

\subsection{Results and Analysis}
The evaluation results for MNIST, CIFAR10, and EuroSAT are shown in Table \ref{classify}. All FL methods achieve the highest performance on MNIST, as this dataset is comparatively easier to learn due to its lower data dimensionality and simpler patterns. MNIST’s grayscale images of handwritten digits are less complex than the color images found in CIFAR10 and EuroSAT. Even in the non-IID setting, where the number of classes per client is limited, the test accuracy only drops by around 1.5\% for all evaluated methods. This suggests that the inherent simplicity of MNIST allows models to generalise well even when trained on a restricted subset of the data.

Using the results on MNIST as a baseline, we observe greater performance drop on CIFAR10 and EuroSAT, largely due to the higher complexity of these datasets, which consist of color images and more intricate patterns. CIFAR10 presents greater variability within its classes, making it more challenging for the FL algorithms to generalise. In the non-IID scenario, FedOpt significantly outperforms the other methods on CIFAR10, demonstrating its robustness in handling heterogeneous data. One of the key differences between FedOpt and other algorithms is that FedOpt incorporates non-linear model aggregation at the server side, as opposed to simply averaging local model updates, which is the case with FedAvg and FedProx. This non-linearity in the global model update process may effectively reconcile the differences between local models, especially when dealing with diverse and non-IID data. The ability of FedOpt to adaptively adjust the aggregation of local models could explain why it is able to capture more complex features in CIFAR10, leading to better performance.

\begin{figure*}[ht]
\centering
\includegraphics[width=0.8\linewidth]{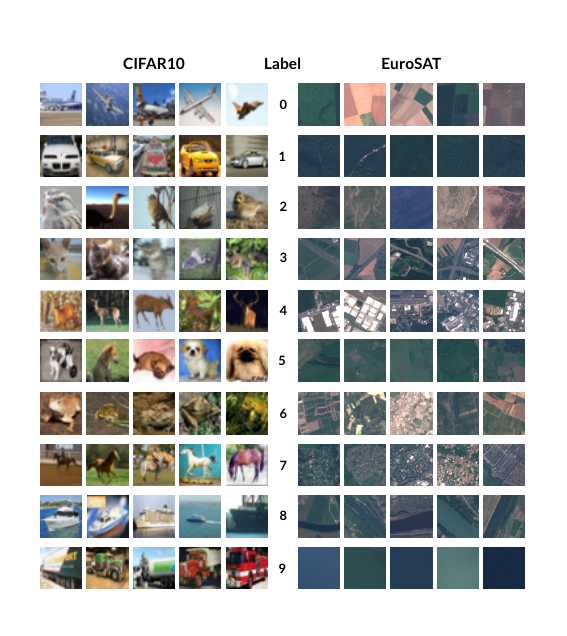}
\caption{Intraclass Variance Comparison between CIFAR10 and EuroSAT. Five randomly selected images from each class illustrate the differences in intraclass variance. EuroSAT images within the same class display higher uniformity in colour and texture, while CIFAR10 images exhibit greater diversity.}
\label{intra}
\end{figure*}

The test accuracy on EuroSAT is higher than on CIFAR10, despite both datasets using the same CNN architecture for their experiments. We attribute this to the lower intra-class variability in the EuroSAT dataset. As shown in Figure \ref{intra}, EuroSAT’s classes exhibit less variation within each class compared to CIFAR10, where objects within the same category can appear significantly different. This lower variability within EuroSAT classes enables the models to achieve higher accuracy, even in the presence of non-IID data. The results suggest that the type of data, rather than just the complexity of the model, can greatly influence the effectiveness of federated learning in UAV swarm applications.

For the image classification tasks summarised in Table \ref{classify}, FedProx demonstrates consistent and stable performance across all datasets, achieving the best results on both MNIST and EuroSAT and remaining competitive on CIFAR10. This stability can be attributed to the proximal term in FedProx, which regularises local model updates by constraining them to remain closer to the global model. This regularisation prevents drastic deviations between local models, which is particularly beneficial in non-IID settings where data distributions can vary widely between clients. 

\begin{table}[ht]
\caption{Evaluation on the CelebA Dataset}
\centering
\begin{tabular}{ccccc}
                        & \multicolumn{4}{c}{}                                  \\ \hline
\multirow{2}{*}{Method} & \multicolumn{2}{c}{IID} & \multicolumn{2}{c}{Non-IID} \\ \cline{2-5} 
                        & Accuracy     & LFW      & Accuracy       & LFW        \\ \hline
FedAvg                  & 29.67        & 64.31    & 13.23          & 61.36      \\
FedProx & 30.83 & \textbf{66.17} & \textbf{14.25} & 63.43 \\
FedOpt & 26.33 & 65.82 & 10.75 & \textbf{64.03} \\
MOON & \textbf{32.25} & 64.37    & 13.83          & 61.57      \\ \hline
\end{tabular}
\\ \%-accuracy of evaluted federated learning algorithms. 
\\ The best results are highlighted \textbf{in bold}.
\label{face}
\end{table}

The results for experiments on the CelebA dataset are reported in Table \ref{face}. Given that the number of available samples is reduced to approximately 6,000 images in total, the test accuracy is significantly lower compared to the results presented in Table \ref{classify} for MNIST, CIFAR10, and EuroSAT. The non-IID scenario on CelebA is particularly difficult because, with 200 subjects distributed across 50 clients, limiting each UAV client to only 4 subjects means that there is no overlap in class labels between any two UAV nodes. This results in a complete partitioning of the data, where each client is working with a highly isolated subset of the dataset. Unlike CIFAR10 or EuroSAT, where there may still be some overlap in classes or similarity between classes, CelebA’s non-IID setting creates a scenario in which each client has a vastly different set of labels. This makes it harder for the global model to learn generalised features that can be applied across different clients.

As mentioned in the last section, we use standard classification accuracy to assess the model's performance on test images of known subjects, similar to the evaluation of general classification tasks. We also perform the LFW test, where the model is tasked with verifying whether two images belong to the same person. In contrast to the standard classification task, the LFW test involves unseen subjects only.

As shown in Table \ref{face}, MOON performs better in the standard accuracy test, while FedProx achieves higher accuracy in the LFW test. We believe the better performance of MOON in the standard test is due to its contrastive learning strategy, which encourages each client to diversify its feature representations based on features from the previous round. This strategy helps MOON produce more discriminative features for known subjects. However,  models trained through MOON struggle with generalizing to unseen identities, as demonstrated by their lower performance on the LFW test. The features learned by MOON may become overfitted to known subjects, leading to difficulties in transferring knowledge to new face identities.

On the other hand, FedProx achieves better generalisation on the LFW test, likely due to its ability to regularise local model updates, preventing excessive divergence between the local and global models. By ensuring that local updates remain closer to the global model, FedProx helps the model learn more general features that are transferable to new, unseen subjects. This regularization is especially important in non-IID settings like CelebA, where the local data is highly personalised and can easily lead to overfitting.

We also experimented with ArcFace, a state-of-the-art face recognition technique that introduces a margin-based contrastive loss to improve the model’s discriminative capability. However, we observed that the global model struggled to converge when using the ArcFace margin loss in the federated learning setup. This can be attributed to the distributed nature of FL, where ArcFace’s core mechanism of pushing negative sample pairs apart and pulling positive pairs together in a common feature space becomes problematic. In centralised learning, all samples share a unified feature space, but in a federated environment, each client constructs its own feature space without awareness of the other clients' spaces. This results in fragmented and contradictory local feature spaces, which can cause inconsistencies during global model aggregation and hinder convergence. The lack of alignment between clients’ feature spaces makes it difficult for the global model to enforce the margin-based loss required by ArcFace effectively.

\section{CONCLUSIONS}
This study has investigated the performance of federated learning algorithms within UAV swarms across various datasets and conditions, illustrating the unique challenges posed by non-IID data distributions inherent to UAV environments. Among the tested algorithms, FedProx has demonstrated notable capability in mitigating the adverse effects of data heterogeneity, attributed largely to its local update strategy for regularising local model deviations.

Given the resource constraints typical of UAV platforms, it is also imperative for future work on benchmarking federated learning in UAV swarms to evaluate the communication overhead and the consumption associated with local training and inference. Additionally, as a next step, we plan to evaluate more federated learning algorithms \cite{scaffold, fedpa, fedboost, element} to further study the challenges of non-IID data in UAV swarms.

While federated learning avoids the need to upload raw data, existing research has shown that privacy leakage can still occur when attackers target local models and infer sensitive information from the gradients or model updates \cite{fl-review}. Therefore, another important direction for future research is developing robust defence strategies against such vulnerabilities when applying FL in UAV swarms.

Moreover, the challenges encountered when attempting to integrate techniques such as ArcFace highlight opportunities for developing domain-specific adaptations. These adaptations would cater to the distributed nature of collaborative learning in UAV swarms. This opens up a research avenue for exploring how techniques designed in the centralised setting can be tailored to meet the specific needs and limitations of UAV platforms.

\addtolength{\textheight}{-12cm}   





\end{document}